\documentclass[10pt,conference]{IEEEtran_custom}
\usepackage{hyperref}

\usepackage{amsmath,amssymb,amsfonts}
\usepackage{algorithm}
\usepackage{algpseudocode}
\usepackage[font=footnotesize,labelfont=bf]{caption}
\usepackage{subcaption}
\usepackage{enumitem}

\usepackage{graphicx}
\usepackage[giveninits=true,hyperref=true,backref=false,backend=biber,maxbibnames=1,maxcitenames=1,style=numeric,citestyle=numeric,sorting=none,url=false,doi=false,isbn=false]{biblatex}
\usepackage{makecell}
\usepackage{multirow}
\usepackage{siunitx}
\usepackage[export]{adjustbox}
\usepackage{dblfloatfix}
\usepackage{svg}
\usepackage{tabularx}
\usepackage{textcomp}
\usepackage{xcolor}
\usepackage{dblfloatfix}
\usepackage{tikz}
\usepackage{booktabs}

\svgsetup{%
    inkscapeformat=pdf, 
    inkscapelatex=false, 
}

\def\titlefontsize{\fontsize{22.88}{27.456}\selectfont}

\algnewcommand{\LineComment}[1]{\State \(\triangleright\) #1}

\newcommand*{\circled}[2][]{\tikz[baseline=(C.base)]{
    \node[inner sep=0pt] (C) {\vphantom{1g}#2};
    \node[draw, circle, inner sep=0.8pt, yshift=1pt]
        at (C.center) {\vphantom{1g}};}}

\usepackage[acronym]{glossaries} 

\newacronym{pe}{PE}{Performance Estimator}
\newacronym{pss}{PSS}{Phase Sequence Selector}

\newacronym{iot}{IoT}{Internet of Things}
\newacronym{cps}{CPS}{Cyber-Physical System}
\newacronym{ml}{ML}{Machine Learning}
\newacronym{sl}{SL}{Supervised Learning}
\newacronym{rl}{RL}{Reinforcement Learning}

\newacronym{pca}{PCA}{Principal Component Analysis}
\newacronym{mle}{MLE}{Maximum Likelihood Estimation}
\newacronym{ir}{IR}{Intermediate Representation}

\usepackage{xargs}    
\usepackage[colorinlistoftodos,prependcaption,textsize=footnotesize]{todonotes}

\newcommandx{\notesdavid}[2][1=]{\todo[linecolor=blue,backgroundcolor=orange!60,bordercolor=blue,#1]{#2}}

\makeatletter
\newcommand{\printfnsymbol}[1]{%
    \textsuperscript{\@fnsymbol{#1}}%
}
\makeatother

\usepackage{fancyhdr}
\fancypagestyle{firstpage}{
  \fancyhf{}
  \chead{To appear at the 24th IEEE/ACM Design, Automation and Test in Europe (DATE'21) Conference, February, 2021.}
}

\begin{document}

\title{\titlefontsize MLComp: A Methodology for \underline{M}achine \underline{L}earning-based Performance Estimation and Adaptive Selection of Pareto-Optimal \underline{Comp}iler Optimization Sequences\vspace*{-3.5mm}}

\author{\IEEEauthorblockN{Alessio~Colucci$^{1,*}$, D\'avid~Juh\'asz$^{2,*}$, Martin~Mosbeck$^{2}$, Alberto~Marchisio$^1$, Semeen~Rehman$^2$, \\ Manfred~Kreutzer$^3$, G\"unther~Nadbath$^3$, Axel~Jantsch$^2$, Muhammad~Shafique$^{4,1}$}
\IEEEauthorblockA{\textit{$^1$Institute of Computer Engineering, Technische Universit{\"a}t Wien (TUWien), Vienna, Austria}}
\IEEEauthorblockA{\textit{$^2$TU Wien, Christian Doppler Laboratory for Embedded Machine Learning, Vienna, Austria}}
\IEEEauthorblockA{\textit{$^3$ABIX GmbH, Vienna, Austria}}
\IEEEauthorblockA{\textit{$^4$Division of Engineering, New York University Abu Dhabi, UAE}}\vspace*{-4mm}\\
Email: \{alessio.colucci,david.juhasz,martin.mosbeck,alberto.marchisio,semeen.rehman,axel.jantsch\}@tuwien.ac.at,\\
\{mkreutzer,gnadbath\}@a-bix.com,muhammad.shafique@nyu.edu\vspace*{-5mm}\\
}

\maketitle
\thispagestyle{firstpage}

\begin{abstract}
\footnotesize
Embedded systems have proliferated in various consumer and industrial applications with the evolution of Cyber-Physical Systems and the Internet of Things.
These systems are subjected to stringent constraints so that embedded software must be optimized for multiple objectives simultaneously, namely reduced energy consumption, execution time, and code size.
Compilers offer optimization phases to improve these metrics. However, proper selection and ordering of them depends on multiple factors and typically requires expert knowledge.
State-of-the-art optimizers facilitate different platforms and applications case by case, and they are limited by optimizing one metric at a time, as well as requiring a time-consuming adaptation for different targets through dynamic profiling.

To address these problems, we propose the novel \textit{MLComp} methodology, in which optimization phases are sequenced by a Reinforcement Learning-based policy. Training of the policy is supported by Machine Learning-based analytical models for quick performance estimation, thereby drastically reducing the time spent for dynamic profiling.
In our framework, different Machine Learning models are \textit{automatically} tested to choose the best-fitting one.
The trained Performance Estimator model is leveraged to efficiently devise Reinforcement Learning-based multi-objective policies for creating quasi-optimal phase sequences.

Compared to state-of-the-art estimation models, our Performance Estimator model achieves lower relative error ({$<\mathrm{\textbf{2\%}}$}) with up to 50$\times$ faster training time over multiple platforms and application domains. Our Phase Selection Policy improves execution time and energy consumption of a given code by up to 12\% and 6\%, respectively.
The Performance Estimator and the Phase Selection Policy can be trained efficiently for any target platform and application domain.

\vspace*{-2mm}

\end{abstract}


\let\thefootnote\relax\footnote{$^*$These authors contributed equally}
\addtocounter{footnote}{-1}\let\thefootnote\svthefootnote

\vspace*{-3mm}
\section{Introduction}
\vspace*{-1mm}
\label{sec:intro}









The number and complexity of embedded systems are constantly growing~\cite{Atzori2010,Samie2016}.
Recent years saw an advent of the \gls*{iot} and \glspl*{cps}, and their subsequent applications~\cite{Liu2017,Yaqoob2014,Massot2009}.
These systems are tightly resource-constrained, requiring latency-limited real-time operation with a very low power budget.
Software running on them must be optimized and tailored to the specific hardware.


Major optimizing compilers, like LLVM~\cite{Lattner2004} and GCC~\cite{Stallman2020}, provide an ever-increasing number of optimization phases to improve operational characteristics of embedded software.
The phases are applied during compilation in sequence.
Their optimal selection and ordering depend on the program to be compiled and on the target platform, as well as on the final optimization objective.
The value of the objective function must be estimated at compile-time to tune phase sequencing.

Standard phase selection and ordering policies in optimizing compilers~\cite{Lattner2004, Stallman2020} are fixed algorithms that have been tuned for the average case and do not exactly fit to actual use-cases.
State-of-the-art approaches for choosing the optimization phases are based on \gls*{ml}, e.g. \gls*{sl} and \gls*{rl}~\cite{Ashouri2017, Fursin2011, Kulkarni2012, ashouriSurveyCompilerAutotuning2018}, and other adaptive mechanisms~\cite{Blackmore2017, Pallister2015}.
Some approaches ignore phase ordering and deal with phase selection only~\cite{Fursin2011,Ashouri2016}, while the order is important for the quality of the generated code~\cite{Almagor2004}.
Most of the works optimize programs for one specific metric only, like execution time~\cite{Ashouri2017,Kulkarni2012,Blackmore2017,Ashouri2016} or energy consumption~\cite{Pallister2015}.
\textit{The state-of-the-art solutions are typically not generic, i.e.,  they provide good results only in a limited environment and for one specific metric at a time.}
Moreover, these methods gather the required metrics by profiling execution, which is super-expensive in time, and should be replaced by a fast-yet-accurate estimation method to reduce the total adaptation time.

State-of-the-art estimation models can be distinguished as \gls*{ml}-based~\cite{diopPowerModelingHeterogeneous2014,balaprakashAutoMOMMLAutomaticMultiobjective2016,liAccurateEfficientProcessor2009} and formal~\cite{bonaEnergyEstimationOptimization2002,vandensteenAnalyticalProcessorPerformance2016,laurentFunctionalLevelPower2004} ones.
\gls*{ml}-based models tend to use accurate sensors and interfaces to estimate the power, and hence require external modifications. However, these methods focus only on a single metric and employ only a small selection of models with at most 5\% relative error.
Formal models estimate the power using formulas and accurate simulation of switching activity that guarantees high accuracy. However, they require deep knowledge of the internal details of the target platform.



\begin{table}[]
\caption{
Comparison of
\gls*{ml}-based state-of-the-art
phase selection policies.}
\vspace*{-1mm}
\resizebox{\linewidth}{!}{%
\setlength\tabcolsep{1.5pt} 
\begin{tabular}{|c|c|c|c|c|c|c|}
\hline
                                                                                           &                                                       & \multicolumn{3}{c|}{\textbf{Metrics}}                                                                                                                             &                                                                                     &                                                                                       \\ \cline{3-5}
\multirow{-2}{*}{\textbf{Solution}}                                                        & \multirow{-2}{*}{\textbf{Technique}}                  & \textit{\begin{tabular}[c]{@{}c@{}}Execution\\ Time\end{tabular}} & \textit{\begin{tabular}[c]{@{}c@{}}Energy\\ Consumption\end{tabular}} & \textit{\begin{tabular}[c]{@{}c@{}}Code\\ Size\end{tabular}}       & \multirow{-2}{*}{\textbf{\begin{tabular}[c]{@{}c@{}}Phase\\ Ordering\end{tabular}}} & \multirow{-2}{*}{\textbf{\begin{tabular}[c]{@{}c@{}}Dynamic\\ Features\end{tabular}}} \\ \hline
COBAYN \cite{Ashouri2016,Ashouri2014}                                     & \acrshort*{sl}                        & X                                            &                                                  &                                         & No                                        & Profiling                                   \\ \hline
Milepost GCC \cite{Fursin2011}                                            & \acrshort*{sl}                        & X                                            &                                                  & X                                       & No                                        & Profiling                                   \\ \hline
MiCOMP \cite{Ashouri2017}                                                 & \acrshort*{sl}                        & X                                            &                                                  &                                         & Static                                    & Profiling                                   \\ \hline
\cite{Kulkarni2012}                                                       & \acrshort*{rl}                        & X                                            &                                                  &                                         & Dynamic                                   & Profiling                                   \\ \hline
\cite{Ashouri2016a}                                                       & \acrshort*{sl}                        & X                                            &                                                  &                                         & Dynamic                                   & Profiling                                   \\ \hline
{\color[HTML]{FF0000} \textit{\textbf{MLComp (\acrshort*{pss})}}} & {\color[HTML]{FF0000} \textit{\textbf{\acrshort*{rl}}}} & {\color[HTML]{FF0000} \textit{\textbf{X}}}                     & {\color[HTML]{FF0000} \textit{\textbf{X}}}                         & {\color[HTML]{FF0000} \textit{\textbf{X}}}                & {\color[HTML]{FF0000} \textit{\textbf{Dynamic}}}            & {\color[HTML]{FF0000} \textit{\textbf{Prediction}}}           \\ \hline
\end{tabular}
}
\label{tab:pss}
\vspace*{-3.2mm}
\end{table}

\begin{table}[]
\caption{
Comparison of state-of-the-art performance estimators.}
\vspace*{-1mm}
\resizebox{\linewidth}{!}{%
\setlength\tabcolsep{1.5pt} 
\begin{tabular}{|c|c|c|c|c|c|c|c|c|}
\hline
                                                   &                                            &                                                                                       & \multicolumn{4}{c|}{\textbf{Metrics}}                                                                                                                                                                                                                                                   &    &                                                                        \\ \cline{4-7}
\multirow{-2}{*}{\textbf{Solution}}                & \multirow{-2}{*}{\textbf{Automation}}      & \multirow{-2}{*}{\textbf{\begin{tabular}[c]{@{}c@{}}Machine\\ Learning\end{tabular}}} & \textit{\begin{tabular}[c]{@{}c@{}}Execution\\ Time\end{tabular}} & \textit{\begin{tabular}[c]{@{}c@{}}Energy\\ Consumption\end{tabular}} & \textit{\begin{tabular}[c]{@{}c@{}}\# Executed\\ Instructions\end{tabular}} & \textit{\begin{tabular}[c]{@{}c@{}}Average\\ Power\end{tabular}} & \multirow{-2}{*}{\textbf{\begin{tabular}[c]{@{}c@{}}Data\\ Gathering\end{tabular}}} & \multirow{-2}{*}{\textbf{Accuracy}} \\ \hline
\cite{diopPowerModelingHeterogeneous2014}                                               & Limited                                           & Basic                                                                                     &                                                                   &                                                                       &                                                                             & X                                                                & Profiling   & $\thicksim$5\%                                                               \\ \hline
\cite{balaprakashAutoMOMMLAutomaticMultiobjective2016}                                        & Limited                                          & Basic                                                                                     & X                                                                 &                                                                       &                                                                             & X                                                                & Profiling & $\thicksim$5\%                                                                 \\ \hline
\cite{liAccurateEfficientProcessor2009}                                                 & Limited                                           & Basic                                                                                     &                                                                   &                                                                       &                                                                             & X                                                                & Simulation   & $\thicksim$3\%                                                              \\ \hline
\cite{bonaEnergyEstimationOptimization2002}                                               & No                                           &  No                                                                                     &                                                                   &                                                                       &                                                                             & X                                                                & Simulation  & $\thicksim$2\%                                                               \\ \hline
\cite{vandensteenAnalyticalProcessorPerformance2016}                                        & Limited                                           & No                                                                                      & X                                                                 &                                                                       &                                                                             & X                                                                & Profiling & $\thicksim$7\%                                                                 \\ \hline
\cite{laurentFunctionalLevelPower2004}                                            & No                                           & No                                                                                      &                                                                   &                                                                       &                                                                             & X                                                                & Simulation &  $<$5\%                                                              \\ \hline
{\color[HTML]{FF0000} \textit{\textbf{\begin{tabular}[c]{@{}c@{}}MLComp\\(\acrshort*{pe})\end{tabular}}}} & {\color[HTML]{FF0000} \textit{\textbf{Full}}} & {\color[HTML]{FF0000} \textit{\textbf{Advanced}}}                                            & {\color[HTML]{FF0000} \textit{\textbf{X}}}                        & {\color[HTML]{FF0000} \textit{\textbf{X}}}                            & {\color[HTML]{FF0000} \textit{\textbf{X}}}                                  & {\color[HTML]{FF0000} \textit{\textbf{X}}}                       & {\color[HTML]{FF0000} \textit{\textbf{Profiling}}} &  {\color[HTML]{FF0000} \textit{\textbf{$<$2\%}}}                      \\ \hline
\end{tabular}%
}
\label{tab:pe}
\vspace*{-5mm}
\end{table}


Our work aims at overcoming specific limitations of state-of-the-art solutions in compiler phase sequencing and performance estimation. Tables~\ref{tab:pss} and~\ref{tab:pe} highlight the shortcomings of major state-of-the-art with respect to different properties/features, and show comprehensive and superior coverage of our \textit{MLComp} methodology.
Our phase sequence selector possesses the following key attributes:
\begin{itemize}[leftmargin=*]
    \item It utilizes \textit{\gls*{rl}}, which has been given little attention in the literature for obtaining optimal phase sequences.
    \item It optimizes programs for multiple objectives in contrast to typical single-objective optimizations.
    \item It supports fast adaptation for different application domains.
\end{itemize}
The latter is enabled by fast performance estimation in the adaptation phase, which requires proper performance modeling of target platforms. To adapt to the different target platforms efficiently, potential models need to be evaluated and the best-fitting one selected. This process, which is usually done by manual analysis and design~\cite{laurentFunctionalLevelPower2004}, is automated in our solution.


Our efforts are focused on the following \textit{scientific challenges}, which, to the best of our knowledge, have not been addressed in the literature before:
\begin{itemize}[leftmargin=*]
    \item automatic evaluation of different \gls*{ml}-based performance models to support adaptation to different platforms;
    \item efficient training of adaptive phase sequence selection policies for multi-objective optimization of programs.
\end{itemize}






\textbf{Novel Contributions:} To address the above challenges, we propose a novel methodology \textit{MLComp} that employs:
\begin{itemize}[leftmargin=*]
    \item \textbf{adaptive analytical models} for estimating energy consumption and execution time, which are trained on 
    features of target applications executing on a given target platform; and
    \item \textbf{adaptive phase sequence selection policies}, which can be trained for quasi-Pareto-optimal code size, energy consumption, and execution time of target applications running on a target platform.
\end{itemize}

The paper makes the following additional \emph{novel contributions}:
\begin{itemize}[leftmargin=*]
    \item \textbf{testing environment} to collect static and dynamic features of target applications on a target platform;
    \item \textbf{framework to adapt performance models} by analyzing and modeling energy consumption and execution time based on the collected code features; and
    \item \textbf{training framework} for \gls*{rl}-based adaptation of phase selection policies with respect to estimated dynamic features, and utilizing trained policies in LLVM.
\end{itemize}

After presenting a brief overview of the required background knowledge in Section \ref{sec:background}, we explain our novel \textit{MLComp} methodology in Section \ref{sec:methodology}. Experimental setup is explained in Section \ref{sec:experimental_setup}, which is followed by our results in Section \ref{sec:evaluation}. Conclusion is drawn in Section \ref{sec:conclusion}.

\vspace*{-1mm}
\section{Background}
\label{sec:background}
\begin{figure}[t]
    \centering
    \includesvg[width=\linewidth]{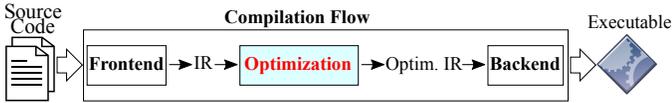}
    \caption{A compiler converts source code into an executable by passing it through optimizations in \acrlong*{ir}.}
    \label{fig:compiler_flow}
    \vspace*{-6mm}
\end{figure}


A compiler converts a given software code implemented in a high-level programming language into machine executable code.
The compilation flow is divided into 3 main parts:
\begin{enumerate}[leftmargin=*]
    \item the \emph{front end} transforms source code into the compiler's \gls*{ir};
    \item the \emph{middle end} performs analyses and transformations in \gls*{ir} to ensure quality and prepares for code generation;
    \item the \emph{back end} generates a target-specific executable from \gls*{ir}.
\end{enumerate}
Optimizing transformations are performed in each stage. Optimizations in the front end are specific to the programming language, while those in the back end tune hardware-specific low-level details. We work with \gls*{ir}-level optimizations in the middle end as depicted in Fig.~\ref{fig:compiler_flow}. Those optimizations are general and applied independently to the source language and target platform. However, they can affect the performance in different ways depending on the target, which calls for their adaptive selection and ordering.


Standard \gls*{ml} methods are available to solve different kinds of algorithmic and modeling problems.
\gls*{sl}~\cite{Simeone2017} is used to learn a model representation that can fit input data to output predictions.
The model is trained in iterative passes. In each pass, the model predicts output for the given input data, and the results are used to update the model weights to return predictions closer to the correct ones.
\gls*{rl}~\cite{Francois-lavet2018}, on the other hand, uses a reward-based system to learn the operations which should be done from the current state of the system itself. The reward reflects how an operation contributes to reaching the objective.



Programs are represented by their characterising features for \gls*{ml} approaches.
The main types of program features are:
\begin{itemize}[leftmargin=*]
    \item \emph{source code features} that characterize application code in a programming language and \gls*{ir} during compilation;
    \item \emph{graph-based features} that provide information about data and control dependencies during compilation;
    \item \emph{dynamic features} that describe operational aspects in architecture-dependent or architecture-independent ways.
\end{itemize}



Architecture-dependent dynamic features such as execution time and energy consumption are objectives for compiler optimization, while static features describe programs during compilation~\cite{Allen1970,Schneck1973,Ferrante1984}.
Dynamic features are time-expensive to determine because they require direct execution of compiled programs. This problem can be circumvented through model-based prediction.
The performance of \gls*{ml} models is improved by scaling and filtering the input features~\cite{ashouriSurveyCompilerAutotuning2018,Wang2018}, an example of which is \gls*{pca}~\cite{Härdle2015,Minka2000}.

\section{The MLComp Methodology}
\label{sec:methodology}
\begin{figure*}[t]
    \centering
    \includesvg[width=\linewidth]{images/methodology_v3.svg}
    \caption{The \textit{MLComp} methodology trains and utilizes \acrlong*{pe} and \acrlong*{pss} in four steps.}
    \label{fig:methodology}
    \vspace*{-5mm}
\end{figure*}


The flow of our methodology is depicted in Fig.~\ref{fig:methodology} and discussed hereafter.
The concept is based on two models:
\begin{enumerate}[leftmargin=*]
    \item \textit{\gls*{pe}} is a fast and efficient way of estimating dynamic features for a given application domain, which is represented by a set of \emph{target applications}, and for a given \emph{target platform}. It allows for accurate prediction adapted to the given domain faster than standard estimation methods.
    \item \textit{\gls*{pss}} is based on a policy for selection and ordering of optimization phases, and supports \gls*{rl}-based adaptation for different \emph{target applications} and \emph{target platforms}. Deploying \gls*{pss} reduces development cost and realizes faster time-to-market by relieving performance engineering from the details of phase selection and ordering.
\end{enumerate}

\subsection{Data Extraction}


\emph{Data Extraction} is the first step of the methodology. We collect a training dataset for the \gls*{pe} model using the flow depicted in box~\circled{1} in Fig.~\ref{fig:methodology}.
Data is extracted for a target platform from a set of target applications by exploring different permutations of optimization phases. Permutations increase the number of data points as differently optimized variants of programs.
For each combination of permutations and applications, the corresponding optimized code is compiled and its features are collected.
We extract \gls*{ir} features that are similar to those of Milepost GCC \cite{Fursin2011}, such as \gls*{ir} instruction counts, data and control dependencies, loop hierarchies, and call chains. Our tool also extracts platform-specific instruction counts from generated code for \gls*{pe} training.
Dynamic features such as execution time, energy consumption, and code size are obtained via profiling the compiled code.
All features are collected in a dataset that is used  for training the \gls*{pe} model. The size of the dataset depends on the specific set of optimizations and the target applications.

\subsection{Performance Estimator (PE)}

\begin{figure}[h]
\vspace*{-6mm}
\begin{algorithm}[H]
\begin{footnotesize}
\captionsetup{font=footnotesize}
\caption{Model search for fitting the \acrlong*{pe} model}
\label{alg:pe_model_search}
\begin{algorithmic}[1]
\vspace*{-2mm}
\Procedure{ModelSearch}{$input$, $accuracy_{thr}$, $list_{models}$}
\LineComment{We initialize $accuracy_{best}$ to the worst case value, which is $-\infty$}
\LineComment{Higher $accuracy$ is better}
\LineComment{$model_{best}$ is initialized to a dummy one, returning a random value}
\State init $accuracy_{best}$, $model_{best}$
\LineComment{split $input$ into $training$ and $testing$ data}
\State $training$, $testing$ $\gets$ split($input$)
\LineComment{We cycle through all the models}
\For{$model$ in $list_{models}$} \label{alg:pe_model_search_for_loop}
\LineComment{After training the $model$, we test it and check its accuracy}
    \State train($model$, $training$)
    \State $accuracy$ $\gets$ test($model$, $testing$)
    \If{$accuracy$ $>$ $accuracy_{best}$}
        \State $accuracy_{best}$ $\gets$ $accuracy$
        \State $model_{best}$ $\gets$ $model$
    \EndIf
    \If{$accuracy_{best}$ $>$ $accuracy_{thr}$}
        \State \textbf{break for loop}
    \EndIf
\EndFor
\State \textbf{return} $model_{best}$, $accuracy_{best}$
\EndProcedure
\vspace*{0mm}
\end{algorithmic}
\end{footnotesize}
\end{algorithm}
\vspace*{-6mm}
\end{figure}


The next step is the \emph{\acrlong*{pe} Model Training} in box~\circled{2} in Fig.~\ref{fig:methodology}.
We search for the preprocessing method and \gls*{ml} model that fits the best to the profiling data based on the code features. The list of methods and models to search is given as input. The search process is detailed in Alg.~\ref{alg:pe_model_search}.
Tables~\ref{tab:pe_preprocessing} and~\ref{tab:pe_models} present the preprocessing methods and \gls*{ml} models used by our \gls*{pe} modeling in this paper.
The set of output metrics is completely customizable.
As a training dataset is collected for one target platform, the \gls*{pe} model is to be trained for each target platform separately to achieve high accuracy.
The trained \gls*{pe} model is used in later steps to predict a program's dynamic features from its \gls*{ir} features.

\subsection{Phase Selection Policy}

\begin{figure}[h]
\vspace*{-5mm}
\begin{algorithm}[H]
\begin{footnotesize}
\captionsetup{font=footnotesize}
\caption{Training the Phase Selection Policy}
\label{alg:pss_training}
\begin{algorithmic}[1]
\vspace*{-2mm}
\Procedure{TrainPolicy}{$programs$, $num\_episodes$, $batch\_size$, $learning\_rate$}
\LineComment{Initialize $policy$ to a random one}
\State init $policy$
\LineComment{Perform training episodes in batches}
\State $episode\_count$ $\gets$ $0$
\While{$episode\_count < num\_episodes$}
    \LineComment{Run episodes and then update $policy$}
    \State $list_{episodes}$ $\gets$ init($batch\_size$, $programs$)
    \State $list_{results}$ $\gets$ run($list_{episodes}$, $policy$)
    \State $policy$ $\gets$ optimize($policy$, $learning\_rate$, $list_{results}$)
    \State $episode\_count$ $\gets$ $episode\_count + batch\_size$
\EndWhile
\State \textbf{return} $policy$
\EndProcedure
\vspace*{0mm}
\end{algorithmic}
\end{footnotesize}
\end{algorithm}
\vspace*{-7mm}
\end{figure}


The trained \gls*{pe} model is used for the \emph{Phase Selection Policy Training} in box~\circled{3} in Fig.~\ref{fig:methodology}.
We use \gls*{rl} to train the policy that selects the best optimization phase to apply to a program characterized by its \gls*{ir} features, and thereby enables an efficient phase sequence to be created iteratively.
The training is done in batches of episodes as listed in Alg.~\ref{alg:pss_training}. The policy is optimized using the REINFORCE \emph{policy gradient method}~\cite{Williams1992,Sutton1999}. The training algorithm creates a phase sequence for a randomly selected target application in each episode with the current policy as depicted in box~\circled{3} in Fig.~\ref{fig:methodology}. The reward in each iteration of an episode reflects how well the last phase changed the dynamic features. Furthermore, the reward guides the training to Pareto-optimal outcomes by penalizing any degradation of the dynamic features. Accumulating rewards over an episode gives a discounted reward, which indicates the overall fitness of the policy for creating a Pareto-optimal phase sequence with respect to final dynamic features. At the end of each batch, the policy is updated according to the episodes' discounted rewards and corresponding phase sequences.
The policy is trained with a given \gls*{pe} and a set of programs that represent a target platform and an application domain, respectively.
The training time is reduced compared to other methods by using \gls*{pe} for fast estimation of dynamic features.
Phase Selection Policy is the model used in the \gls*{pss}.

\subsection{Phase Sequence Selection (PSS)}


The last step of \textit{MLComp} is \emph{Deployment} in box~\circled{4} in Fig.~\ref{fig:methodology}, which is the \gls*{pss} utilizing a trained Phase Selection Policy.
We apply the \gls*{pss} model to drive a compiler's optimizer by selecting phases one after the other.
The policy predicts how probable it is that a phase improves dynamic features of the program and accordingly the phase with the highest probability is applied.
In case the selected phase did not change the program, which might happen because of the uncertainty of the selection, the best predicted phase remains the same for the next iteration. \gls*{pss} overcomes that situation by applying the second best, the third best, and so on until a predefined limit, which is ``Max. inactive subsequence length'' in Table~\ref{tab:pss_training_parameters}.
Phase selection ends when that limit is reached or when the total number of applied phases reaches a threshold.

Note, \gls*{pss} does not require a \gls*{pe} model because the policy learns the platform-specific knowledge.
Decoupling the \gls*{pe} and \gls*{pss} models allows their separate training so that a platform vendor might provide a trained \gls*{pe} model for application developers, who can train a custom \gls*{pss} model with a set of representative applications.

Although the \gls*{pss} model is trained to reach the Pareto-front by selecting locally optimal phases, we observe that both the \gls*{pe} and \gls*{pss} models have \emph{approximation uncertainties} and true Pareto-optimality can not be guaranteed.
The accuracy of \gls*{pss} might be quantified by applying \emph{probabilistic dominance}~\cite{Khosravi2019}, which requires an in-depth empirical evaluation and statistical analysis beyond the scope of this paper. Our evaluation in Section~\ref{sec:evaluation} still shows quasi-Pareto-optimality of the results.

\section{Experimental Setup}
\label{sec:experimental_setup}
\begin{figure}[t]
    \centering
    \includesvg[width=\linewidth]{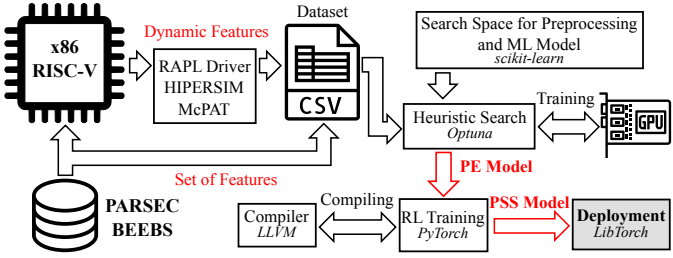}
    \vspace*{-1mm}
    \caption{Detailed toolflow for \textit{MLComp} experimental setup. The process is repeated for each pair of target platform and applications, and it can be adapted for other pairs with minimal code changes.}
    \label{fig:experimental_setup}
\end{figure}

\begin{table}[]
\footnotesize
\centering
\caption{Preprocessing algorithms evaluated during \gls*{pe} training. The list can easily be expanded, as our framework is customizable with different libraries.}
\begin{tabular}{|c|c|c|}
\hline
\multicolumn{3}{|c|}{\textbf{Preprocessing Algorithms}}                \\ \hline
PCA              & Kernel PCA        & NCA                  \\ \hline
Mean-Std Scaling & Min-Max Scaling   & Max-Abs Scaling      \\ \hline
Robust Scaling   & Power Transformer & Quantile Transformer \\ \hline
\end{tabular}%
\label{tab:pe_preprocessing}
\vspace*{-3mm}
\end{table}

\begin{table}[]
\footnotesize
\centering
\caption{\gls*{ml} models evaluated during \gls*{pe} training. The list can easily be expanded, as our framework is customizable with different libraries.}
\setlength\tabcolsep{1.5pt} 
\begin{tabular}{|c|c|c|}
\hline
\multicolumn{3}{|c|}{\textbf{Machine Learning Regression Models}}                                       \\ \hline
Ridge                     & Kernel Ridge                 & Bayesian Ridge                   \\ \hline
Linear         & SGD               & Passive-Aggressive    \\ \hline
ARD            & Huber             & Theil-Sen             \\ \hline
LARS                      & Lasso                        & Lasso-LARS                       \\ \hline
Support Vector & Nu-Support Vector & Linear Support Vector \\ \hline
ElasticNet                & Orthogonal Matching Pursuit   & Multi-Layer Perceptron           \\ \hline
Decision Tree             & Extra Tree        & Random Forest                    \\ \hline
\end{tabular}
\label{tab:pe_models}
\vspace*{-2mm}
\end{table}

\begin{table}[]
\vspace*{-4mm}
\caption{Parameters of \gls*{pss} training.}
\resizebox{\linewidth}{!}{%
\begin{tabular}{|l|r|l|r|}
\hline
\multicolumn{1}{|c|}{\textbf{Parameter}} & \multicolumn{1}{|c|}{\textbf{Value}} & \multicolumn{1}{|c|}{\textbf{Parameter}} & \multicolumn{1}{|c|}{\textbf{Value}} \\ \hline
Number of layers & $3$ & Size of inner layer & $16$ \\ \hline
Number of episodes & $512$ & Batch size & $6$ \\ \hline
Max. phase sequence length & $128$ & Learning Rate & $0.1$  \\ \hline
Max. inactive subsequence length & $8$ & & \\ \hline
\end{tabular}%
}
\label{tab:pss_training_parameters}
\vspace*{-2mm}
\end{table}

\begin{table}[]
\footnotesize
\centering
\caption{LLVM optimization phases used for \gls*{pss} evaluation. The list can easily be expanded. These phases are from optimization levels \texttt{-O3} and \texttt{-Oz}.}
\setlength\tabcolsep{1.5pt} 
\begin{tabular}{|c|c|c|}
\hline
\multicolumn{3}{|c|}{\textbf{Optimization Phases}} \\ \hline
adce & aggressive-instcombine & alignment-from-assumptions \\ \hline
argpromotion & bdce & called-value-propagation \\ \hline
callsite-splitting & constmerge & correlated-propagation \\ \hline
deadargelim & div-rem-pairs & dse \\ \hline
early-cse & early-cse-memssa & elim-avail-extern \\ \hline
float2int & globaldce & globalopt \\ \hline
globals-aa & gvn & indvars \\ \hline
inline & instcombine & instsimplify \\ \hline
ipsccp & jump-threading & licm \\ \hline
loop-deletion & loop-distribute & loop-idiom \\ \hline
loop-load-elim & loop-rotate & loop-sink \\ \hline
loop-unroll & loop-unswitch & loop-vectorize \\ \hline
lower-expect & mem2reg & memcpyopt \\ \hline
mldst-motion & prune-eh & reassociate \\ \hline
sccp & simplifycfg & slp-vectorizer \\ \hline
speculative-execution & sroa & tailcallelim \\ \hline
\end{tabular}%
\label{tab:pss_phases}
\vspace*{-4mm}
\end{table}


A detailed toolflow for our experimental setup is shown in Fig. \ref{fig:experimental_setup}. We worked with two different target platforms: profiling for an \emph{x86 target} is done on an Intel Core i7 system using the RAPL~\cite{davidRAPLMemoryPower2010} interface to measure power consumption, and dynamic features for a \emph{RISC-V target} are obtained by accurate simulation with the industrial-grade simulator HIPERSIM~\cite{hipersim} integrated with the open-source McPAT~\cite{Sheng2009}.
Programs are compiled with LLVM~\cite{Lattner2004} version 9.0.0, which is able to target both platforms. The size of the collected dataset depends on the target applications and the set of optimization phases chosen at compile time: in this evaluation, we used between 200 and 600 data points for both the PARSEC benchmark \cite{Bienia2008} on \textit{x86 target} and the BEEBS benchmark \cite{Pallister2013} on \textit{RISC-V target}.

The training of the \gls*{pe} model is implemented in Python using Optuna~\cite{optuna_2019}, scikit-learn~\cite{scikit-learn}, and pandas~\cite{rebackPandasdevPandasPandas2020}. It covers preprocessing methods and \gls*{ml} models listed in Tables~\ref{tab:pe_preprocessing} and~\ref{tab:pe_models}, respectively. The set of output metrics has been chosen to analyze different patterns and distributions. Even though power consumption has a slight correlation with execution time, as in the number of cycles and the number of instructions~\cite{Pallister2015}, increasing the complexity of the system reduces the correlation of these metrics. Therefore, each of them is important for learning the dynamic behaviour of the system.

\gls*{pss} training is implemented also in Python using PyTorch~\cite{paszkePyTorchImperativeStyle2019} for realizing the model. The 63 code features that our static analysis obtains are preprocessed by \gls*{pca} with \gls*{mle}~\cite{Minka2000} before being passed to Deep \gls*{rl}~\cite{Francois-lavet2018}. The \gls*{pss} model is trained with the parameters listed in Table~\ref{tab:pss_training_parameters}, applying optimization phases shown in Table~\ref{tab:pss_phases}.
The trained model is stored in TorchScript format, to be loaded into and utilized by our custom LLVM optimization with LibTorch, the PyTorch C++ API.

Note that \gls*{pe} and \gls*{pss} are independent of the target platform and the used application set. The necessary adaptations when changing the target platform are limited to adjusting target-specific compiler flags and utilizing a tool with support for gathering dynamic features inside the \emph{Data Extraction} block. Furthermore, any application can be used with our training frameworks as long as it supports a build method using LLVM and allows controlling optimization phases via parameters or environment variables.

\vspace*{-2mm}
\section{Evaluation}
\label{sec:evaluation}
To evaluate our \textit{MLComp} methodology, we trained and tested both \gls*{pe} and \gls*{pss} models on different target platforms with different benchmarks as target applications.

\subsection{PARSEC Benchmark Evaluation on x86 Platform}

\begin{figure}[t]
    \centering
    \includesvg[width=\linewidth]{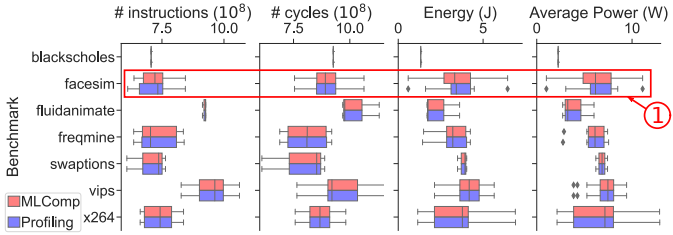}
    \caption{Comparison between profiling data and prediction of a trained \gls*{pe} model for PARSEC benchmark applications on x86 platform shows very similar distributions, which supports the efficacy and accuracy of our model.}
    \label{fig:pe_x86_parsec}
    \vspace*{-3mm}
\end{figure}

Here we focus our analysis on the PARSEC benchmark \cite{Bienia2008}, running on an x86 platform. First, we gather the required dataset by profiling the execution of programs from the benchmark compiled with different optimization phases. Then, we use our framework to train different \gls*{ml} models and select the best one; the results are shown in Fig. \ref{fig:pe_x86_parsec}. As we can see, the exact distributions and the ones generated by our \gls*{pe} model are almost identical for all the 4 metrics.
Note that the \textit{blackscholes} benchmark has a very tight distribution, while all the others have wider distributions. Referring to \circled{1}, we can see that the only visible difference resides in the \textit{facesim} benchmark. However, there is always a high fidelity, as the error between the correct and the predicted distributions always has the same bias. This property is important for the training of the \gls*{pss} model, giving the correct positive/negative reward to the current choice, even if a limited prediction error is present.

\begin{figure}[t]
    \centering
    \includesvg[width=\linewidth]{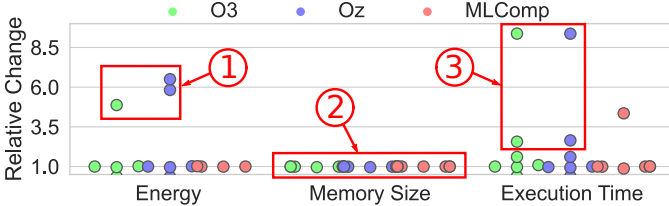}
    \caption{\gls*{pss} validation results for PARSEC applications on x86 platform. Values are relative to those of unoptimized code, the lower is the better.}
    \label{fig:pss_x86_parsec}
    \vspace*{-2mm}
\end{figure}

After validating the \gls*{pe} model, we used it to train the corresponding \gls*{pss} model. In Fig. \ref{fig:pss_x86_parsec}, we can see the result of the validation executed after the training. Specifically, distributions are pretty similar across standard state-of-the-art optimizations and \textit{MLComp}. However, in some cases, as shown by \circled{1} and \circled{3}, some standard phase usage can increase both the energy consumption and the execution time between 8x and 10x, respectively, while \textit{MLComp} shows slight improvements. Regarding memory size, as pointed by \circled{2}, there are minimal gains, which could be related to the benchmarks being synthetic applications.

\subsection{BEEBS Benchmark Evaluation on RISC-V Platform}

\begin{figure}[t]
    \centering
    \includesvg[width=\linewidth]{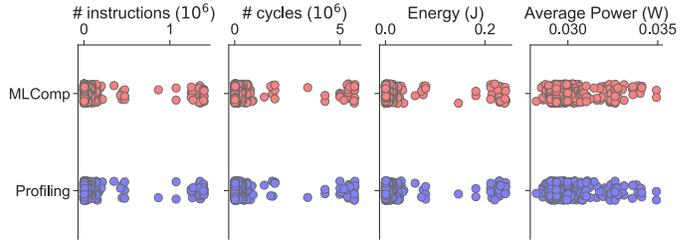}
    \caption{Comparison between profiling data and prediction of a trained \gls*{pe} model for BEEBS benchmark applications on RISC-V platform.}
    \label{fig:pe_riscv_beebs}
    \vspace*{-3mm}
\end{figure}

We performed a similar evaluation for BEEBS \cite{Pallister2013} on the RISC-V platform. In this case, the number of benchmarks is much higher compared to PARSEC, and since the \gls*{pe} results are similar to those with PARSEC, we show only an overview of the distribution points in Fig. \ref{fig:pe_riscv_beebs}.

\begin{figure}[t]
    \centering
    \includesvg[width=\linewidth]{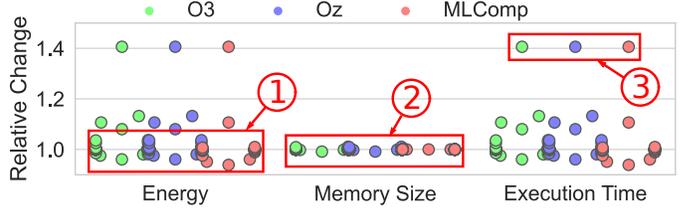}
    \caption{\gls*{pss} validation results for BEEBS applications on RISC-V platform. Values are relative to those of unoptimized code, the lower is the better.}
    \label{fig:pss_riscv_beebs}
    \vspace*{-4mm}
\end{figure}

This \gls*{pe} model was then used to train a \gls*{pss} model, obtaining the results shown in Fig. \ref{fig:pss_riscv_beebs}. Here, at pointer \circled{1}, we can see that our \textit{MLComp} performs better on average than standard state-of-the-art policies: reducing energy while also optimizing other objectives. Also with BEEBS, we can see that the memory size does not improve or worsen much, as pointed by \circled{2}. In addition, \textit{MLComp} results in similar patterns of execution time and energy consumption. Focusing on pointer \circled{3}, we can see how our \textit{MLComp} obtain more balanced results compared to standard state-of-the-art policies.

\subsection{Discussion and Key Takeaways}

Our \gls*{pe} model has a maximum percentage error \textit{smaller than 2\%} across all four metrics, while that of the comparable state of the art is in the range of 2\%-7\% on a single metric \cite{diopPowerModelingHeterogeneous2014, balaprakashAutoMOMMLAutomaticMultiobjective2016,liAccurateEfficientProcessor2009,bonaEnergyEstimationOptimization2002,vandensteenAnalyticalProcessorPerformance2016,laurentFunctionalLevelPower2004}.
Moreover, the efficient setup for data extraction and the heuristic search of models help us to reach higher accuracy with less time spent for acquiring data and training the model. Profiling the applications and training the models took \textit{only 2 days}, compared to 15, 30 or 108 days \cite{laurentFunctionalLevelPower2004, bonaEnergyEstimationOptimization2002}.

Drawing a straight comparison is more difficult for our \gls*{pss} model, as related techniques optimize a single objective only.
The state-of-the-art results oscillate between 5\% and 30\% improvement in execution time~\cite{Ashouri2017,Ashouri2016a}, which makes our results fall in their average with up to 12\% improvement in that metric.
However, our \gls*{pss} model considers additional objectives and reaches up to 6\% reduction in energy consumption while not increasing code size. There is actually a slight 0.1\% improvement in the latter.

We can summarize the following key observations:

\begin{itemize}[leftmargin=*]
    \item The \gls*{pe} model realizes fast estimation with high accuracy, as it is capable of reproducing the profiled distributions.
    \item The \gls*{pss} model performs better than standard optimizations on average and also provides quasi-optimal results for multiple objectives.
    \item Our \textit{MLComp} methodology is fully automated and is usable with different target platforms and applications, enabling for fast estimation and optimization without manual analysis and modeling required.
\end{itemize}

\vspace*{-1.5mm}
\section{Conclusion}
\label{sec:conclusion}


We propose the \textit{MLComp} methodology to overcome limitations of current solutions in compiler optimization phase sequencing and performance modeling.
State-of-the-art optimizers can be applied for different target platforms and applications case by case, but their adaptation is expensive and they typically optimize one metric only.
\textit{MLComp} supports adaptive selection of Pareto-optimal phase sequences with respect to execution time, power consumption, and code size by a \acrfull*{pss} with an \gls*{rl}-based policy. Fast adaptation of the policy for different target platforms and application domains is enabled by an \gls*{ml}-based \acrfull*{pe} model, which provides fast-yet-accurate prediction of dynamic program features. The \gls*{pe} model is trained for a target platform by automatically selecting the most suitable data preprocessing method and \gls*{ml} model for accurate prediction. This is a novel contribution in performance modeling as current solutions require manual analysis and modeling.
Experiments with LLVM on the x86 and RISC-V platforms show that our methodology is efficiently reusable with different target platforms and applications. The \gls*{pe} model realizes fast estimation with very high accuracy, and the \gls*{pss} model performs better than state-of-the-art optimizations with multiple objectives.

\vspace*{-1mm}
\section*{Acknowledgment}


The presented work has been conducted in the ``Cost Efficient Smart System Software Synthesis - COGUTS II (Code Generation for Ultra-Thin Systems)'' project, funded and supported by the Austrian Research Promotion Agency (FFG) under grant agreement 872663, and affiliated under the umbrella of the EU Eureka R\&D\&I ITEA3 ``COMPACT'' Cluster programme.

\vspace*{-2mm}
\printbibliography


\end{document}